\def\year{2022}\relax
\documentclass[letterpaper]{article} 
\usepackage{aaai22}  
\usepackage{times}  
\usepackage{helvet}  
\usepackage{courier}  
\usepackage[hyphens]{url}  
\usepackage{graphicx} 
\usepackage{natbib}  
\usepackage{caption} 
\DeclareCaptionStyle{ruled}{labelfont=normalfont,labelsep=colon,strut=off} 
\usepackage{algorithm}
\usepackage{algorithmic}

\usepackage{newfloat}
\usepackage{listings}
\usepackage{amsmath}
\usepackage{amsfonts}
\usepackage{amssymb}
\usepackage{adjustbox}
\usepackage{makecell}
\usepackage{multirow}
\usepackage{xcolor}
\usepackage{textcomp}
\usepackage{longtable}

\floatstyle{ruled}
\newfloat{listing}{tb}{lst}{}
\floatname{listing}{Listing}
\title{MINIMAL: Mining \textit{Models} for Universal Adversarial Triggers}
\author{ Yaman Kumar Singla\thanks{Equal Contribution} \textsuperscript{\rm 1,3,5},
Swapnil Parekh$^*$ \textsuperscript{\rm 2},
Somesh Singh$^*$ \textsuperscript{\rm 3}, \\
Balaji Krishnamurthy \textsuperscript{\rm 1},
Rajiv Ratn Shah \textsuperscript{\rm 3}, 
Changyou Chen\textsuperscript{\rm 5} 
       
}
\affiliations{
\textsuperscript{\rm 1}Adobe Media Data Science Research, 
\textsuperscript{\rm 3}IIIT-Delhi,
\textsuperscript{\rm 5}SUNY at Buffalo\\
\textsuperscript{\rm 2}New York University,
}

\usepackage{bibentry}

\newcommand{\mb}[1]{\boldsymbol{\mathbf{#1}}}
\newcommand{\loss}{\ensuremath\mathcal{L}}
\newcommand{\PreserveBackslash}[1]{\let\temp=\\#1\let\\=\temp}
\newcolumntype{C}[1]{>{\PreserveBackslash\centering}p{#1}}
\newcolumntype{R}[1]{>{\PreserveBackslash\raggedleft}p{#1}}
\newcolumntype{L}[1]{>{\PreserveBackslash\raggedright}p{#1}}
  
\definecolor{adversarial}{rgb}{0.90, 0.02, 0.03}
\definecolor{orange2}{rgb}{0.95,0.35,0}
\definecolor{trigger}{HTML}{FFC7BF}

\newcommand{\cy}[1]{}
\DeclareMathOperator*{\argmax}{arg\,max}
\DeclareMathOperator*{\argmin}{arg\,min}

\begin{document}

\maketitle

\vspace*{-10mm}

\begin{abstract}
It is well known that natural language models are vulnerable to adversarial attacks, which are mostly input-specific in nature. Recently, it has been shown that there also exist input-agnostic attacks in NLP models, special text sequences called universal adversarial triggers. However, existing methods to craft universal triggers are data intensive. They require large amounts of data samples to generate adversarial triggers, which are typically inaccessible by attackers. For instance, previous works take 3000 data samples per class for the SNLI dataset to generate adversarial triggers. In this paper, we present a novel data-free approach, \textit{MINIMAL}, to mine input-agnostic adversarial triggers from models. Using the triggers produced with our data-free algorithm, we reduce the accuracy of Stanford Sentiment Treebank's positive class from 93.6\% to 9.6\%. Similarly, for the Stanford Natural Language Inference (SNLI), our single-word trigger reduces the accuracy of the entailment class from 90.95\% to less than 0.6\%. Despite being completely data-free, we get equivalent accuracy drops as data-dependent methods\footnote{The code and reproducibility steps are given in \url{https://anonymous.4open.science/r/data-free-uats-B9B1}}. 
\end{abstract}


\section{Introduction}
\label{sec:introduction}
In the past two decades, deep learning models have shown impressive performance over many natural language tasks, including sentiment analysis \cite{zhang2018deep}, natural language inference \cite{parikh2016decomposable}, automatic essay scoring \cite{kumar2019get}, question-answering \cite{xiong2016dynamic}, keyphrase extraction \cite{meng2017deep}, \textit{etc}. At the same time, it has also been shown that these models are highly vulnerable to adversarial perturbations \cite{behjati2019universal}. The adversaries change the inputs to cause the models to make errors. Adversarial examples pose a significant challenge to the rising deployment of deep learning based systems.

Commonly, adversarial examples are found on a per-sample basis, \textit{i.e.}, a separate optimization needs to be performed for each sample to generate an adversarially perturbed sample. Since the optimization needs to be performed for each sample, it is computationally expensive and requires deep learning expertise for generation and testing. Lately, several research studies have shown the existence of input-agnostic universal adversarial trigger (UATs) \cite{moosavi2017universal,wallace2019universal}. These are a sequence of tokens, which, when added to any example, cause a targeted change in the prediction of a neural network. The existence of such word sequences poses a considerable security challenge since the word sequences can be easily distributed and can cause a model to predict incorrectly for all of its inputs. Moreover, unlike input-dependent adversarial examples, no model access is required at the run time for generating UATs. At the same time, the analysis of universal adversaries is interesting from the point of view of model, dataset analysis and interpretability (\S\ref{sec:Analysing the Class Impressions}). They tell us about the global model behaviour and the general input-output patterns learnt by a model \cite{wallace2019universal}.

Existing approaches to generate UATs assume that an attacker can obtain the training data on which a targeted model is trained \cite{wallace2019universal,behjati2019universal}. While generating an adversarial trigger, an attacker firstly \textit{trains} a proxy model on the training data and then generates adversarial examples by using gradient information. Table~\ref{table:NSamples} presents the data requirements during training for the current approaches. For instance, to find universal adversaries on the natural language inference task, one needs 9000 training examples. Also, the adversarial ability of a perturbation has been shown to depend on the amount of data available \cite{mopuri2017fast,mopuri2018generalizable}. However, in practice, an attacker rarely has access to the training data. Training data are usually private and hidden inside a company's data storage facility, while only the trained model is publicly accessible. For instance, Google Cloud Natural Language (GCNL) API only outputs the scores for the sentiment classes \cite{gcnlAPI} while the data on which the GCNL model was trained is kept private. In this real-world setting, most of the adversarial attacks fail.

\begin{figure}
 \centering
 \begin{tabular}{c}
\hspace{-0.3cm}\includegraphics[scale=0.45]{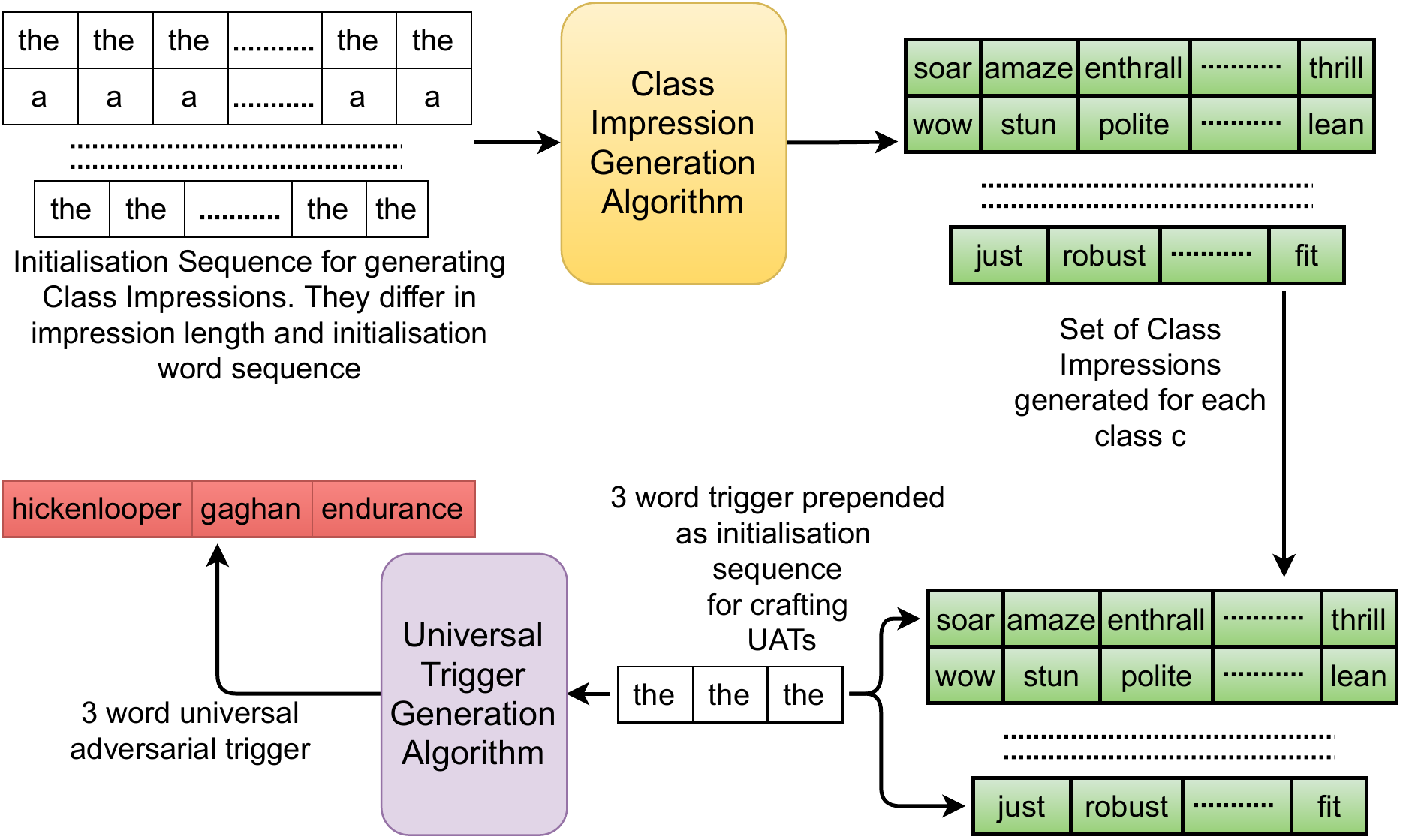}
 \end{tabular}

 \caption{\small 
 Two step process to generate universal adversarial triggers. First, we generate multiple class impressions for each class $c$. For this, we take multiple initialization sequences differing in starting word and length. After generating class impressions, we use them as our dataset for generating universal adversarial triggers.}

 \label{fig:2 stage ci uat}
\end{figure}

In this paper, we present a novel data-free approach for crafting universal adversarial triggers to address the above issues. Our method is to mine a trained \textit{model} (but not data) for perturbations that can fool the target model without any knowledge about the data
distribution (\textit{e.g.}, type of data, length and vocabulary of samples, \textit{etc}.). We only need access to the embedding layer and model outputs. Our method achieves this by solving first-order Taylor approximation of two tasks: first, we generate ``class-impressions'' (\S\ref{sec:Class-Impressions Generation Algorithm}), which are reconstructed text sentences from a model's memory representing the learned parameters for a certain data class; and second, we mine universal adversarial triggers over these generated class impressions (\S\ref{sec:Trigger Generation Algorithm}). Class-impression can be considered as the general representation of samples belonging to a particular class (Fig~\ref{fig:t-SNE plots}) and are used to emulate samples belonging to that class in our method. The concept of data leaving its impression on a trained model has also been observed in prior work in model inversion attacks in computer vision \cite{micaelli2019zero,nayak2019zero}. We build on that concept to mine universal adversarial triggers. We propose a combination of general model inversion attacks methodology with trigger generation to mine data-free adversarial triggers and show our results for several NLP models \cite{fredrikson2015model,tramer2016stealing}.

\noindent The major contributions of our work are summarized as:

\noindent- For the first time in the literature, we propose a novel data-free approach, MINIMAL (MINIng Models for AdversariaL triggers), to craft universal adversarial triggers for natural language processing models and achieve state-of-the-art success (adversarial) rates (\S\ref{sec:experiments}). We show the efficacy of the triggers generated using our method on three well-known datasets and tasks, \textit{viz.}, sentiment analysis (\S\ref{sec:sentiment analysis}) on Stanford Sentiment Treebank (SST) \cite{socher2013recursive}, natural language inference (\S\ref{sec:Natural Language Inference}) on the SNLI dataset \cite{bowman2015large}, and paraphrase detection (\S\ref{sec:Paraphrase Identification}) on the MRPC dataset \cite{dolan2005automatically}.

\noindent- We use both class impressions and universal adversarial triggers generated by our models to try to understand the models' global behaviour (\S\ref{sec:Analysing the Class Impressions}). We observe that the words with the lowest entropy (\textit{i.e.}, the most informative features) appear in the class impressions (Fig.~\ref{fig:entropy-all-3-datasets}). We find that these low entropy word-level features can also act as universal adversarial triggers (Table~\ref{table:snli-uat-ci-word-transfer}). The class-impression words are good representations of a class since they form distinct clusters in the manifold representations of each class.

\section{Related Work}
\label{sec:related work}

\textbf{Universal Adversarial Attacks:} \citet{moosavi2017universal} showed the existence of universal adversarial perturbations. They showed that a \textit{single perturbation} could fool DNNs most of the times when added to all images. Since then, many universal adversarial attacks have been designed for vision systems \cite{khrulkov2018art,li2019universal,zhang2021survey}. To the best of our knowledge, there are only three recent papers for NLP based universal adversarial attacks, and all of them require data for generating universal adversarial triggers \cite{wallace2019universal,song2020universal,behjati2019universal}. In simultaneous works, \cite{wallace2019universal,behjati2019universal} show universal adversarial triggers for NLP. \citet{song2020universal} extend it to generate natural (data-distribution like) triggers. We compare our work with \cite{wallace2019universal} since they show improved adversarial success rates over \cite{behjati2019universal}. We leave mining natural triggers from models as a future study. Our results demonstrate comparable performance as \cite{wallace2019universal} but without using any data. Table~\ref{table:NSamples} mentions the data requirement of \cite{wallace2019universal}.

\begin{figure}[htbp]
 \centering
 \begin{tabular}{c}
\includegraphics[scale=0.52]{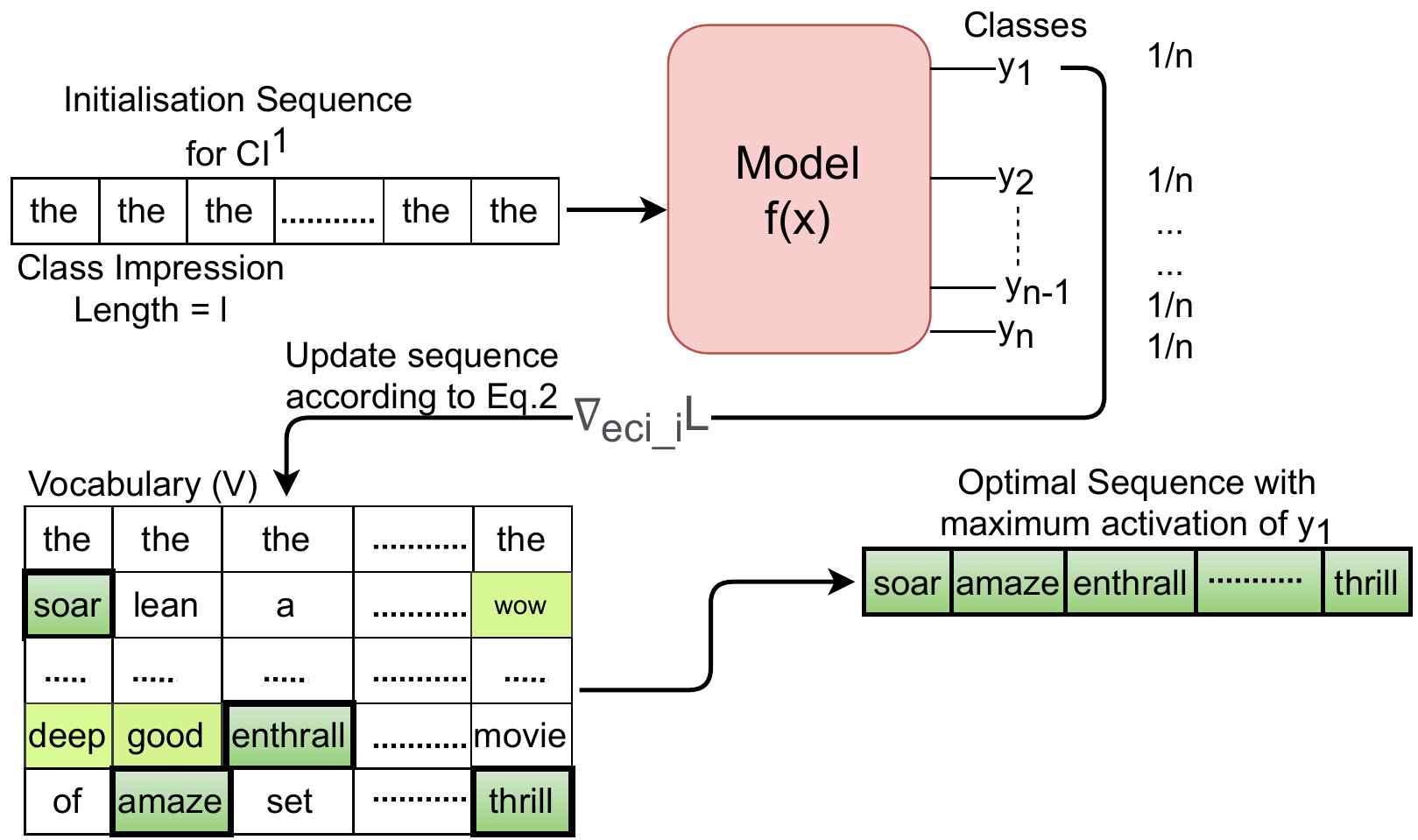}
 \end{tabular}

 \caption{\small 
 Class Impression Generation (CIG) Algorithm. We start with an initial sequence of \textit{``the the ... the''} and continuously update it based on its gradient with respect to output probabilities (Eq.~\ref{eq:class impressions taylor}). The final sequence we get represents the class impression $CI^c$ for the class $c$.
 }

 \label{fig:Class Impression Generation Algorithm}
\end{figure}

While there are many proposed classifications of adversarial attacks, from the point of view of our work, they can be seen in two ways: (a)~data-based attacks; (b)~data-free attacks. Data-based computer vision attacks depend on training and validation dataset to craft adversaries, while data-free attacks rely on other signals. There are some data-free approaches in computer vision, for example, by maximizing activations at each layer \cite{mopuri2017fast,mopuri2018generalizable}, class activations \cite{mopuri2018ask}, and pretrained models and proxy dataset \cite{huan2020data}. However, there has been no work in NLP systems for data-free attacks.

\section{The Proposed Approach}
\label{sec:proposed approach}
In summary, our algorithm of crafting data-free universal adversarial triggers is divided into two steps, as shown in Fig~\ref{fig:2 stage ci uat}. First, we generate a set of class-impressions (\S\ref{sec:Class-Impressions Generation Algorithm}) (Fig~\ref{fig:Class Impression Generation Algorithm}) for each class. These natural language examples represent the entire class of samples and are generated solely from the weights learnt by the model. Second, we use the set of class impressions generated in the first step to craft universal adversarial triggers corresponding to those impressions (\S\ref{sec:Trigger Generation Algorithm}) (Fig~\ref{fig:UAT Generation Algorithm}).

\subsection{Class-Impressions Generation (CIG) Algorithm}
\label{sec:Class-Impressions Generation Algorithm}

To generate the class impression $CI^c$ for a class $c$, we propose to maximize the confidence of the model $f(x)$ for an input text sequence $\mb{t}_c$. Formally, we maximize:
\begin{equation}
\label{eq:class impression optimization}
CI^c = \argmax_{\mb{t}_{c}} \mathbb{E}_{\mb{t}_{c} \sim \mathcal{V}}\left[\loss(c, f(\mb{t}_c)) \right],
\end{equation}
\noindent where $\mb{t}_c$ is sampled from a vocabulary $\mathcal{V}$. The input $\mb{t}_c$ in NLP is not continuous, but is made up of discrete tokens. Therefore, we use the first-order Taylor approximation of Eq.~\ref{eq:class impression optimization} \cite{michel2019evaluation,ebrahimi2017hotflip,wallace2019universal}. Formally, for every token $\mb{e}_{ci_i}$ in a class impression $CI^c$, we solve the following equation: 
\begin{equation}\label{eq:class impressions taylor}
 \mb{e}_{ci_i} = \argmin_{\mb{e}_i^\prime \in \mathcal V}\left[\mb{e}_i^\prime-{\mb{e}_{ci_i}}\right]^\intercal\nabla_{\mb{e}_{ci_i}}\loss,
\end{equation}
\noindent where $\mathcal{V}$ represents the set of all words in the vocabulary, and $\nabla_{\mb{e}_{ci_i}}\loss$ is the gradient of the task loss. We model the Eq.~\ref{eq:class impressions taylor} as an iterative procedure by starting out with an initialisation value of $\mb{e}_{ci_i}$ as \textit{`the'}. We then continually optimize it until convergence. For computing the optimal $\mb{e}_i^\prime$, we take $\vert \mathcal V \vert$ $d$-dimensional dot products where $d$ is the dimensionality of the token embedding. We use beam-search for finding the optimal sequence of tokens $\mb{e}_i^\prime$ to get the minimum loss in Eq.~\ref{eq:class impressions taylor}. We score each beam using the loss on the batch in each iteration of the optimization schedule.

Finally, we convert the optimal $\mb{e}_{ci_i}$ back to their associated word tokens. Fig.~\ref{fig:Class Impression Generation Algorithm} presents an overview of the process. It shows the case where we initialized $\mb{e}_{ci_i}$ with a sequence of \textit{``the the .... the''} and then follow the optimization procedure for finding the optimal $CI^c$ for the class $c$\footnote{We vary the initialization sequence and sequence length to generate multiple class impressions for the same class}.

To generate class impressions for the models that use contextualized embeddings like BERT \cite{devlin2018bert}, we perform the above optimization over character and sub-word level. We also replace the context-independent embeddings in Eq.~\ref{eq:class impressions taylor} with contextual embeddings as obtained from BERT after passing the complete sentence to it.

We generate multiple class impressions for each class for all models by varying the number of tokens and the starting sequence. This gives us a number of class impressions for the next step where we generate triggers over these class impressions.

\begin{table}
{\begin{tabular}{|l||l||l|l|l|}
\hline
\textbf{Dataset} & \textbf{\makecell{Validation Size\\(Real samples)}} & \textbf{\makecell{Impressions Size\\(Generated samples)}} \\ \hline
  SST & 900 & 300 \\ \hline
  SNLI & 9000 & 400 \\ \hline
  MRPC & 800 & 300 \\ \hline
  
\end{tabular}}
\caption{\label{table:NSamples}
\small
Number of Samples required to generate Universal Adversarial Triggers for each Dataset. In a data-based approach like \cite{wallace2019universal}, validation set (column 2) is used to generate the UATs. The third column lists the number of queries we make to generate artificial samples. These artificial samples are then used to craft UATs. Note that no real samples are required for our method.}
\end{table}

\subsection{The Universal Trigger Generation (UTG) Algorithm}
\label{sec:Trigger Generation Algorithm}

\begin{figure}[htbp]
 \centering
 \begin{tabular}{c}
\includegraphics[scale=0.45]{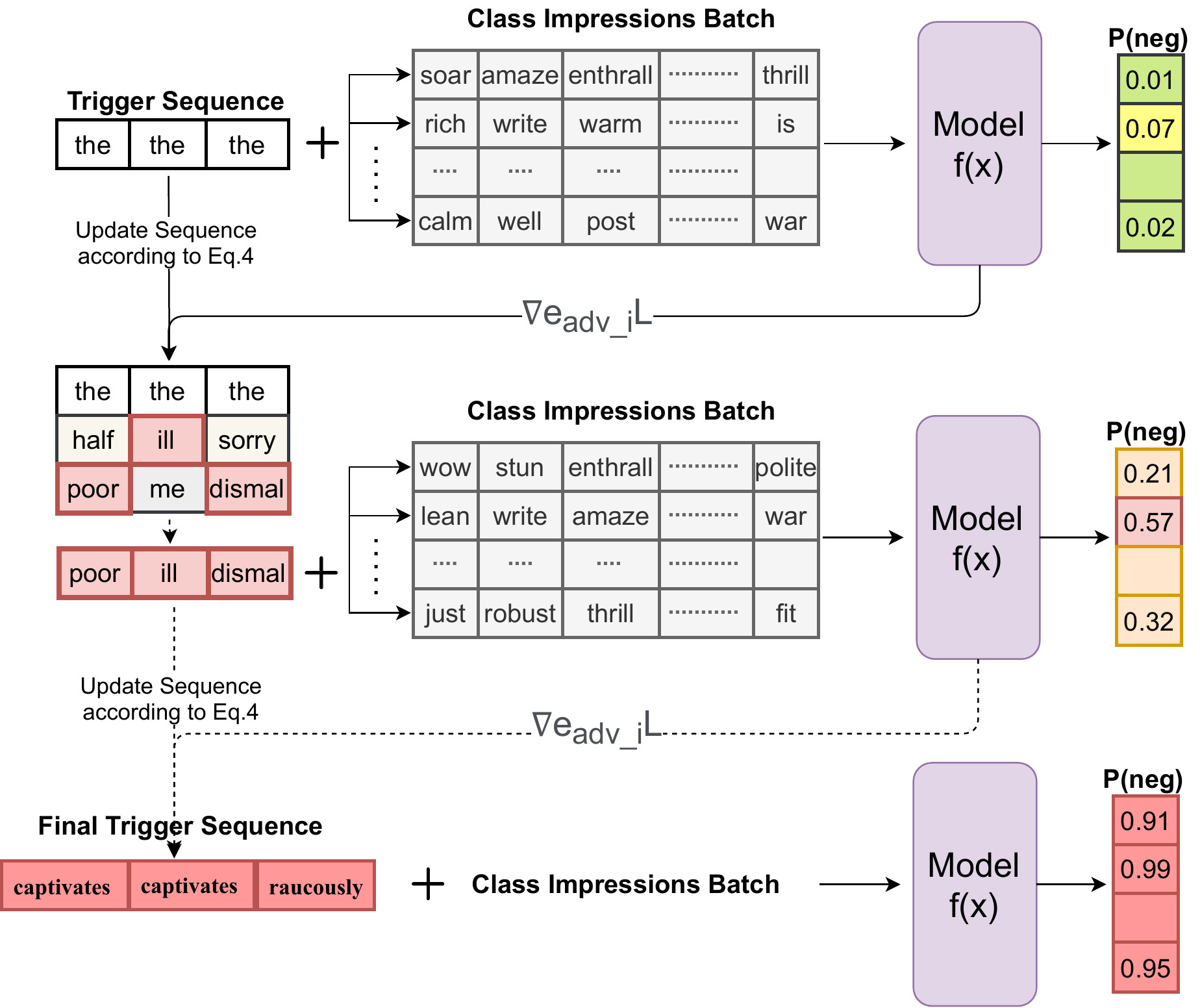}
 \end{tabular}

 \caption{\small Iterative Universal Trigger Generation (UTG) Algorithm.
 }
 \label{fig:UAT Generation Algorithm}
\end{figure}

\begin{table*}[!htbp]
{
\begin{adjustbox}{width=\textwidth}

\begin{tabular}{|l||l|}
\hline
\textbf{Class} & \textbf{Class Impression} \\ \hline
  {\bf Positive} & energizes energizes captivated energizes enthrall eye-catching captivating aptitude artistry passion\\ \hline
  {\bf Positive} & captures soul-stirring captivates mesmerizing soar amaze excite amaze enthrall thrill captivating impress artistry accomplishments\\ \hline
  {\bf Negative} & spiritless ill-constructed ill-conceived ill-fitting aborted fearing bottom-rung woe-is-me uncharismatically pileup\\ \hline
  {\bf Negative} & laziest third-rate insignificance stultifyingly untalented hat-in-hand rot leanest blame direct-to-video wounds urinates\\ \hline
 
\end{tabular}
\end{adjustbox}

}
\caption{\label{table:SST-GLOVE-CI} \small Class Impressions for BiLSTM-Word2Vec Sentiment Analysis  Model. Note that the words in the class impression examples highly correspond to the respective sentiment classes.}
\end{table*}

After generating class impressions in the previous step, we generate adversarial triggers as follows. From the last algorithm, we get a batch of class impressions $CI^c$ for the class $c$. The task of crafting universal adversarial triggers is defined as minimizing the following loss function:
\begin{equation}\label{eq:uat}
\argmin_{\mb{t}_{adv}} \mathbb{E}_{\mb{t} \sim \mathcal CI^c}\left[\loss(\tilde{c}, f(\mb{t}_{adv}; \mb{t})) \right],
\end{equation}

\noindent where $\tilde{c}$ denotes target class (distinct from the class $c$), $f(\mb{t}_{adv}; \mb{t})$ denotes the evaluation of $f(x)$ on the input containing concatenation of adversarial trigger tokens at the start of the text $t$. The text $t$ is sampled from the set of all class impressions $CI^c$. Again, we use the Taylor approximation of the above equation. Therefore, we get:
\begin{equation}\label{eq:uat taylor}
 \mb{e}_{adv_i} = \argmin_{\mb{e}_i^\prime \in \mathcal V}\left[\mb{e}_i^\prime-{\mb{e}_{adv_i}}\right]^\intercal\nabla_{\mb{e}_{adv_i}}\loss,
\end{equation}

\noindent where $\mathcal{V}$ represents the set of all words in the vocabulary, and $\nabla_{\mb{e}_{adv_i}}\loss$ is the average gradient of the task loss over a batch. We model Eq.~\ref{eq:uat taylor} as an iterative procedure where we initialize $\mb{e}_{adv_i}$ with an initialisation value of `the'. For computing the optimal $\mb{e}_i^\prime$, similar to the previous step, we take $\vert \mathcal V \vert$ $d$-dimensional dot products where $d$ is the dimensionality of the token embedding. We use beam-search for finding the optimal sequence of tokens $\mb{e}_i^\prime$ to get the minimum loss in Eq.~\ref{eq:uat taylor}. We score each beam using the loss on the batch in each iteration of the optimization schedule.
Additionally, to generate impressions of varying difficulty, we randomly select the token from a N-sized beam of possible minimal candidates, instead of the least scoring candidate.
\begin{table}[!h]
{
\begin{adjustbox}{width=\columnwidth}
\begin{tabular}{|l||l||l|l|l|}
\hline
\textbf{Type} & \textbf{Direction} & \textbf{Trigger} & \textbf{\makecell{Acc.\\Before}} &\textbf{\makecell{Acc.\\After}} \\ \hline
  Data-based & P $\rightarrow$ N & \makecell{worthless\\endurance useless} & 93.6 & 9.6 \\ \hline
  Data-free & P $\rightarrow$ N  & \makecell{useless\\endurance useless} & 93.6 & 9.6 \\ \hline
  Data-based & N $\rightarrow$ P & \makecell{kid-empowerment\\hickenlooper enjoyable} & 80.3 & 7.9 \\ \hline
  Data-free & N $\rightarrow$ P  & \makecell{compassionately\\hickenlooper gaghan} & 80.3 & 8.1 \\ \hline
  
\end{tabular}
\end{adjustbox}
}
\caption{\label{table:SST-GLOVE} \small
The table reports the accuracy drop for the BiLSTM-Word2Vec sentiment analysis model after prepending 3-word adversarial triggers generated using MINIMAL and data-based methods.}
\end{table}

\begin{table}[!h]
{
\begin{adjustbox}{width=0.9\columnwidth}
\begin{tabular}{|l||l||l|l|l|}
\hline
\textbf{Type} & \textbf{Direction} & \textbf{Trigger} & \textbf{\makecell{Acc.\\Before}} &\textbf{\makecell{Acc.\\After}} \\ \hline
  Data-free & P $\rightarrow$ N  & \makecell{useless\\endurance useless} & 86.2 & 32 \\ \hline
  \hline
  Data-free & N $\rightarrow$ P  & \makecell{compassionately\\hickenlooper gaghan} & 86.9 & 35 \\ \hline
 
\end{tabular}
\end{adjustbox}
}

\caption{\label{table:SST-transfer-ELMO} \small Accuracy drop for transfer attack with data-free UAT generated by our method. We prepend 3-word adversarial triggers to the SST BiLSTM-ELMo model.}
\end{table}

\begin{table}
\scalebox{0.98}{
\begin{adjustbox}{width=0.9\columnwidth}

\begin{tabular}{|c||c|}
\hline
\textbf{Class} & \textbf{Class Impressions} \\ \hline

{\bf Contradiction} & \makecell{\textbf{Hypothesis}: lynched cardinals giraffes lynched\\lynched a mown extremist natgeo illustration\\\textbf{Premise}: zucchini restrooms swimming golds\\weekday rock 4 seven named dart} \\ \hline
 
{\bf Entailment} & \makecell{\textbf{Hypothesis}: civilization va physical supersonic\\prohibits biathlon body land muffler mobility\\\textbf{Premise}: gecko robed abroad teetotalers blonds\\pluggling sprinter speeds corks dogtrack} \\ \hline

{\bf Neutral} & \makecell{\textbf{Hypothesis}: porters festivals fluent a playgrounds\\ratatouille buttercups horseback popularity waist\\\textbf{Premise}: bowler teaspoons group tourism tourism\\spiritual physical physical  person} \\ \hline
\end{tabular}
\end{adjustbox}
}
\caption{\label{table:SNLI-GLOVE-CI}\small Class Impressions for ESIM model trained for the Natural Language Inference Task}


\footnotesize{
\begin{adjustbox}{width=0.88\columnwidth}
\begin{tabular}{|c|c|c|c||c|}
\hline

\multicolumn{5}{|c|}{\makecell{\textbf{Class Type: Entailment}$\rightarrow${\textbf{Neutral}}\\Original Accuracy: (ESIM: 91\%, DA: 90.3\%)}}\\\hline
\makecell{\textbf{Data-Inputs}} & \textbf{Data-Type} & \textbf{Trigger} & \textbf{ESIM} & \textbf{DA} \\ \hline

\multirow{2}{*}{\makecell{\\\textbf{Hypothesis and}\\\textbf{Premise}}}
& Data-Based & \makecell{nobody\\whatsoever\\cats} & \makecell{0.06\\0.6\\0.69} &  \makecell{0.18\\43\\0.7}\\\cline{2-5}
 & Data-Free& \makecell{nobody\\no\\mars} &  \makecell{0.06\\0.1\\0.1} &  \makecell{0.18\\2\\0.3}\\ \hline

\textbf{Hypothesis-Only} & Data-Free & \makecell{monkeys\\zebras\\cats} & \makecell{0.7\\0.5\\0.69} &  \makecell{0.54\\0.39\\0.7}\\ \hline
\hline

\multicolumn{5}{|c|}{\makecell{\textbf{Class Type: Neutral}$\rightarrow${\textbf{Contradiction}}\\Original Accuracy: (ESIM: 88\%, DA: 80\%)}}\\\hline
\makecell{\textbf{Data-Inputs}} & \textbf{Data-Type} & \textbf{Trigger} & \textbf{ESIM} & \textbf{DA} \\ \hline

\multirow{2}{*}{\makecell{\\\textbf{Hypothesis and}\\\textbf{Premise}}}& Data-Based &\makecell{shark\\moon\\spacecraft} & \makecell{18\\17\\12} &  \makecell{28\\13\\8.4}\\ \cline{2-5}

&Data-Free &\makecell{skydiving\\orangutan\\spacecraft} & \makecell{14\\12\\12} &  \makecell{20\\75\\8.4} \\ \cline{1-5}

\textbf{Hypothesis-Only}&Data-Free &\makecell{sleep \\drowning\\spacecraft } & \makecell{11\\15\\12} &  \makecell{19\\29\\8.4} \\ \hline
\hline

\multicolumn{5}{|c|}{\makecell{\textbf{Class Type: Contradiction}$\rightarrow${\textbf{Entailment}}\\Original Accuracy: (ESIM: 79\%, DA: 85\%)}}\\\hline
\makecell{\textbf{Data-Inputs}} & \textbf{Data-Type} & \textbf{Trigger} & \textbf{ESIM} & \textbf{DA} \\ \hline

\multirow{2}{*}{\makecell{\\\textbf{Hypothesis and}\\\textbf{Premise}}}&Data-Based & \makecell{expert\\siblings\\championship} & \makecell{64\\66\\65} &  \makecell{73\\68\\74}\\
\cline{2-5}
& Data-Free&\makecell{inanimate\\final\\championships} & \makecell{67\\66\\68} &  \makecell{82\\68\\85} \\ 
\hline

\textbf{Hypothesis-Only}& Data-Free &\makecell{humans\\semifinals\\championship} & \makecell{70\\68\\65} &  \makecell{79\\74\\74} \\ \hline
\hline

\end{tabular}
\end{adjustbox}
}
\caption{\small We prepend a single word (Trigger) to SNLI hypotheses. We display the top 3 triggers created using both Validation set and Class Impressions for ESIM and show their performance on the DA. The original accuracies are mentioned in brackets. \cy{Hard to compare. Why not add another row of average accuracy?}}
\label{table:snli-uat}

\scalebox{0.98}{
\begin{adjustbox}{width=0.9\columnwidth}
\begin{tabular}{|c||c|}
\hline
\textbf{Class} & \textbf{Class Impressions} \\ \hline

{\bf \makecell{Paraphase\\Detected}}     & \makecell{\textbf{Sentence 1}:nintendo daredevil bamba bamba the\\the lakers dodgers weekend rhapsody seahawks\\\textbf{Sentence 2}:  nintendo multiplayer shawnee dodgers \\  anthem netball the olympics soundtrack\\overture martial} \\ \hline
 \hline
{\bf \makecell{Paraphase\\Detected}}     & \makecell{\textbf{Sentence 1}: mon submitted icus submit arboretum\\templar desires them requirements kum \\\textbf{Sentence 2}: lection rahul organizers postgraduate \\ qualifying your exercises signifies its them} \\ \hline
\hline
{\bf \makecell{No Paraphase\\Detected}} & \makecell{\textbf{Sentence 1}: b 617 matrices dhabi ein wm spelt\\rox a proportional alamo swap \\\textbf{Sentence 2}:  drilled traced 03 02 said\\mattered million 0\% 50\% corporations a a}\\ \hline
 \hline
{\bf \makecell{No Paraphase\\Detected}} & \makecell{\textbf{Sentence 1}: cw an hung kanda singapore\\tribu chun mid 199798 nies bula latvia\\\textbf{Sentence 2}: came tempered paced times than\\an saying say shone say s copp}\\\hline
\hline
\end{tabular}
\end{adjustbox}
}
\caption{\label{table:MRPC-ALBERT-CI} Class Impressions for ALBERT model trained for the Microsoft Research Paraphrase Corpus \small}
\end{table}

Finally, we convert the optimal $\mb{e}_{adv_i}$ back to their associated word tokens. Fig.~\ref{fig:UAT Generation Algorithm} presents an overview of the process. Similar to Sec.~\ref{sec:Class-Impressions Generation Algorithm}, we initialize the iterative algorithm with a sequence ($\mb{e}_{adv}$) of \textit{``the the .... the''}\footnote{We vary the initialisation sequence and sequence length to generate multiple adversarial triggers} and then follow the optimization procedure to find the optimal $\mb{e}_{adv}$. We handle contextual embeddings in a similar manner as in Sec.~\ref{sec:Class-Impressions Generation Algorithm}. Next, we show the application of the algorithms developed on several downstream tasks.
\section{Experiments}
\label{sec:experiments}
We present our experimental setup and the effectiveness of the proposed method in terms of the success rates achieved by the crafted UATs. We test our method on several tasks including sentiment analysis, natural language inference, and paraphrase detection.

\subsection{Sentiment Analysis}
\label{sec:sentiment analysis}
We use the Stanford Sentiment Treebank (SST) dataset \cite{socher2013recursive}. Previous studies have extensively used this dataset for studying sentiment analysis \cite{devlin2018bert,cambria2013new}. We use two models on this dataset: Bi-LSTM model \cite{graves2005framewise} with word2vec emebddings \cite{mikolov2017advances}, Bi-LSTM model with ELMo embeddings \cite{peters2018deep}. The same models have been used in previous work \cite{wallace2019universal} for generating data-dependent universal adversarial triggers. The models achieve an accuracy of 84.4\% and 86.6\% over the dataset, respectively. We compare our algorithm with \cite{wallace2019universal} since it is demonstrated to work better than other works \cite{behjati2019universal}.

\noindent\textbf{Class Impressions:} First, we generate class impressions for the model. Table~\ref{table:SST-GLOVE-CI} presents 2 class impressions per class. As can be seen from the table, the words selected by the CIG algorithm highly correspond to the class sentiment. For instance, the algorithm selects positive words such as {\em energizes, enthrall} for the positive class, and negative words such as {\em spiritless, ill-conceived, laziest} for the negative class. We posit that the class impressions generated through our algorithm can be used to interpret what a model has learnt.


\noindent\textbf{UAT:} Next, using the class impressions generated for the models, we generate universal adversarial triggers with the UTG algorithm (Sec~\ref{sec:Trigger Generation Algorithm}). In order to avoid selecting construct-relevant words, we remove such words\footnote{\url{https://www.cs.uic.edu/~liub/FBS/sentiment-analysis.html#lexicon}} from our vocabulary for this task. Table~\ref{table:SST-GLOVE} shows the results for the performance of adversarial triggers generated using our method and those by the data-based approach of \cite{wallace2019universal}. Despite being completely independent of data, we achieve comparable accuracy drops as \cite{wallace2019universal}. We are able to reduce the sentiment prediction accuracy by more than 70\% for both the classes.


\textbf{Transfer of Mined UATs:} We check whether the triggers mined from one model also work on other models. For this, we test the triggers mined from BiLSTM-Word2Vec model on the BiLSTM-ELMo model. Table~\ref{table:SST-transfer-ELMO} notes the results for the same. The triggers reduce the accuracy for both the classes by more than 50\%. This is significant since they are completely mined from the model without any information of the underlying distribution.

\subsection{Natural Language Inference}
\label{sec:Natural Language Inference}
For natural language inference, we use the well-known Stanford Natural Language Inference (SNLI) Corpus \cite{bowman2015large}. We use two models for our analysis on this task: Enhanced Sequential Inference Model (ESIM) \cite{chen2016enhanced} and Decomposable Attention (DA) \cite{parikh2016decomposable} with GloVe embeddings \cite{pennington2014glove}. The accuracies reported by ESIM is 86.2\%, and DA is 85\%.


\noindent\textbf{Class Impressions:} Modelling natural language inference involves taking in two inputs: premise and hypothesis and deciding the relation between them. The relation can be one amongst entailment, contradiction, and neutral. Following the algorithm in Sec.~\ref{sec:Class-Impressions Generation Algorithm}, we find both premise and hypothesis together after starting out from a common initial word sequence. Through this, we get a \emph{typical} premise and its corresponding hypothesis for the three output classes (entailment, contradiction, and neutral). 

One example per class for the ESIM model is given in Table~\ref{table:SNLI-GLOVE-CI}. Unlike sentiment analysis, class impressions for SNLI are not readily interpretable. This is because that while a sentence from the SST corpus can be considered a combination of latent sentiments, the same cannot be assumed of a hypothesis sentence from the SNLI corpus. A statement by itself is not a characteristic hypothesis (or premise). For instance, the SST sentence ``You'll probably love it.'' is a characteristic positive polarity sentence and can be understood to be so by the word `love'. The same cannot be said for the SNLI premise sentence ``An older and younger man smiling.'' SNLI class impressions give us a glance into a model's learnt deep manifold representation of premise-hypothesis pair. They are generally far away from the training data. Strong priors about the natural training distribution might be needed to make them closer to the training data, . We leave this task for future investigation.


\textbf{UAT:} After obtaining a batch of class impressions from the previous step, we craft the universal adversarial triggers. A comparison of the results for UATs generated using our method, and those of \cite{wallace2019universal} are given in Table~\ref{table:snli-uat}. As can be seen, we achieve comparable results as \cite{wallace2019universal}. A single word trigger is able to reduce the accuracy of the entailment class from 90.3\% to 0.06\%.

\textbf{Hypothesis Only UATs:} Several recent research studies have indicated that the annotation protocol for SNLI leaves artefacts in the dataset such that by giving just hypothesis, one can obtain 67\% accuracy \cite{gururangan2018annotation,poliak2018hypothesis}. Following that line of study, we generate only the hypothesis class impressions using the CIG algorithm. Then, we generate triggers over the hypothesis-only generated class impressions. Table~\ref{table:snli-uat} notes the results for the hypothesis-only attacks. We find that hypothesis-only triggers perform equivalently to hypothesis and premise attacks. This provides further proof that there are many biases in the SNLI dataset and more importantly, the models are using those biases as class representations and adversarial triggers actively exploit these (\S\ref{sec:Analysing the Class Impressions}).



\textbf{Transfer of Mined UATs to Other Models:} To determine how the triggers mined from one model transfer to another, we test both data-based and our data-free triggers generated using the ESIM model on the DA model. Table~\ref{table:snli-uat} shows the results. We check the transfer attack performance in two cases: where both hypothesis and premise are given and where only the hypothesis is given. It can be seen that even though both the models are architecturally very different, the triggers transfer remarkably well for both cases. For instance, for the entailment class, the original and transfer attack accuracy drops are comparable. It is also noteworthy that our results are equivalent to \cite{wallace2019universal} even for transfer attacks.

\subsection{Paraphrase Identification}
\label{sec:Paraphrase Identification}
For paraphrase identification, we use the Microsoft Research Paraphrase Corpus (MRPC) \cite{dolan2005automatically}. Paraphrase identification is the task of identifying whether two sentences are semantically equivalent. We use the ALBERT model \cite{lan2019albert} for the task. It reports an accuracy of 89.9\% over this.

\noindent \textbf{Class Impressions:} Similar to natural language inference, here, the models require two input sentences. The task of the model is to identify whether the two sentences are semantically the same. The class impressions generated on the ALBERT model are given in Table~\ref{table:MRPC-ALBERT-CI}. We find that similar to the SNLI corpus, the MRPC class impressions are not readily interpretable. For specific examples like the first example in the table, we find that sometimes words related to one topic occur as class impressions. Words like `nintendo' and 'daredevil' in sentence one and `multiplayer' and `anthem' often occur in the context of multiplayer digital games. We should have got similar class impressions in an ideal scenario for sentences 1 and 2 for actual paraphrases. However, we find that the model considers even those sentence pairs (example 2) as paraphrases that have zero vocabulary or topic overlap. This indicates that the model is performing a similarity match in the high dimensional data manifold. We do some analysis for this in Sec.~\ref{sec:Analysing the Class Impressions}. We leave the further investigation of this for future work.

\textbf{UAT:} Table~\ref{table:MRPC-Albert} notes the performance of 3 word data-free adversarial triggers generated using MINIMAL. As can be seen, the mined artefacts reduce the accuracy for both classes by more than 70\%.

\section{Analyzing the Class Impressions}
\label{sec:Analysing the Class Impressions}
We futher analyze class impressions and their relationship with universal adversarial triggers. Specifically, we try to answer these questions: which words get selected as class impressions, why are we able to find universal adversarial triggers from a batch of class impressions and no train data distribution is required? We also try to relate it to the observation made by \cite{gururangan2018annotation,poliak2018hypothesis}, which ranked the dataset artefact words by calculating their pointwise-mutual information (PMI) values for each class. We further show that the trigger words align very well with dataset artefacts.

\begin{table}[!htbp]
{
\setlength{\tabcolsep}{2pt}
\begin{adjustbox}{width=\columnwidth}

\begin{tabular}{|l||l||l|l|l|}
\hline
\textbf{Type} & \textbf{Direction} & \textbf{Trigger} & \textbf{\makecell{Acc.\\Before}} &\textbf{\makecell{Acc.\\After}} \\ \hline
  Data-free & P $\rightarrow$ N  & \makecell{insisting sacrificing either} & 95 & 45\\ \hline
  Data-free & N $\rightarrow$ P  & \makecell{waistband interests stomped} & 80.9 & 61.6\\ \hline
\end{tabular}
\end{adjustbox}
}
\caption{\label{table:MRPC-Albert} \small Accuracy drop for the ALBERT paraphrase identification model after prepending 3-word adversarial triggers generated using MINIMAL.}

\begin{adjustbox}{width=\columnwidth}

\begin{tabular}{|c|c|c|c|}
\hline
\multicolumn{4}{|c|}{\textbf{Stanford Sentiment Treebank}}          \\ \hline
\textbf{Positive}  & \textbf{\%} & \textbf{Negative}  & \textbf{\%} \\ \hline
beautifully        & 99.97       & dull               & 99.99       \\ \hline
wonderful          & 99.95     & worst              & 99.99       \\ \hline
enjoyable          & 99.94       & suffers            & 99.98       \\ \hline
engrossing         & 99.94       & stupid             & 99.98       \\ \hline
charming           & 99.89       & unfunny            & 99.97       \\ \hline \hline
Impression Average & 73.89       & Impression Average & 77.97       \\ \hline
\end{tabular}
\end{adjustbox}
\caption{\small PMI percentiles for sample class impression words and their average}
\label{table:sst-pmi-percentile}

\begin{adjustbox}{width=\columnwidth}

\begin{tabular}{|c|c|c|c|}
\hline
\multicolumn{4}{|c|}{\textbf{Microsoft Research Paraphrase Corpus}}               \\ \hline
\textbf{Paraphrase}         & \textbf{\%} & \textbf{Non-Paraphrase} & \textbf{\%} \\ \hline
experts                     & 99.89       & biological              & 99.91       \\ \hline
{\color[HTML]{000000} such} & 99.84       & important               & 99.39       \\ \hline
only                        & 99.67       & drug                    & 99.92       \\ \hline
due                         & 99.65       & case                    & 98.91       \\ \hline
said                        & 99.57       & among                   & 98.73       \\ \hline
Impression Average          & 77.23       & Impression Average      & 81.89       \\ \hline
\end{tabular}
\end{adjustbox}
\caption{\small PMI percentiles for sample class impression words and their average}
\label{table:mrpc-pmi-percentile}

\begin{adjustbox}{width=\columnwidth}
\begin{tabular}{|lc|lc|lc|}
\hline
\multicolumn{6}{|c|}{\textbf{Stanford Natural language Inference}}                                          \\ \hline
\textbf{Contradiction} & \textbf{\%} & \textbf{Entailment} & \textbf{\%} & \textbf{Neutral}   & \textbf{\%} \\ \hline
naked                  & 99.99       & human               & 99.91       & about              & 99.73       \\ \hline
sleeping               & 99.97       & athletic            & 99.73       & treasure           & 99.06       \\ \hline
tv                     & 99.96       & martial             & 99.71       & headed             & 99.05       \\ \hline
asleep                 & 99.96       & clothes             & 99.53       & school             & 98.87       \\ \hline
eats                   & 99.93       & aquatic             & 99.38       & league             & 98.83       \\ \hline \hline
Average     & 67.89       & Average  & 70.89       & Average & 68.97       \\ \hline
\end{tabular}
\end{adjustbox}
\caption{\small PMI percentiles for sample class impression words and their average}
\label{table:snli-pmi-percentile}

\centering
\footnotesize{
\begin{adjustbox}{width=\columnwidth}

\begin{tabular}{|c|c|c|}
\hline
\textbf{Ground Truth$\rightarrow$Attacked Target} & \textbf{Trigger} & \textbf{ESIM} \\ \hline

\vtop{\hbox{\strut {\textbf{Entailment}}$\rightarrow${\textbf{Neutral}}}\hbox{\strut Accuracy:88\% }}
& \makecell{beatboxing\\insects\\reclining} & \makecell{77\\68\\83}\\ \hline

\vtop{\hbox{\strut {\textbf{Entailment}}$\rightarrow${\textbf{Contradiction}}}\hbox{\strut Accuracy:79\%}}
& \makecell{qualities\\coexist\\stressful} & \makecell{70\\71\\70}\\ \hline

\vtop{\hbox{\strut {\textbf{Neutral}}$\rightarrow${\textbf{Contradiction}}}\hbox{\strut Accuracy:   79\%}}
& \makecell{disoriented\\arousing\\championship} & \makecell{69\\67\\65} \\ \hline

\vtop{\hbox{\strut {\textbf{Neutral}}$\rightarrow${\textbf{Entailment}}}\hbox{\strut Accuracy:   91\%}}
& \makecell{championship\\semifinals\\aunts} & \makecell{0.1\\0.9\\0.5} \\ \hline

\vtop{\hbox{\strut {\textbf{Contradiction}}$\rightarrow${\textbf{Entailment}}}\hbox{\strut Accuracy: 91\% }}
&\makecell{ballet\\nap\\olives} & \makecell{5\\2\\9}\\ \hline

\vtop{\hbox{\strut {\textbf{Contradiction}}$\rightarrow${\textbf{Neutral}}}\hbox{\strut Accuracy:   88\% }}
& \makecell{nap\\hubble\\snakes} & \makecell{14\\21\\9} \\ \hline

\hline

\end{tabular}
\end{adjustbox}
}
\caption{\small We prepend a single word (trigger) to SNLI hypotheses. We take the first word from all ground truth class impressions and evaluate them on class impressions of the target class. We then choose the top 4 and show their validation performance for the target class.
\label{table:snli-uat-ci-word-transfer}}
\end{table}

\begin{figure*}[htbp]
 \centering
 \begin{tabular}{ccc}
\includegraphics[scale=0.35]{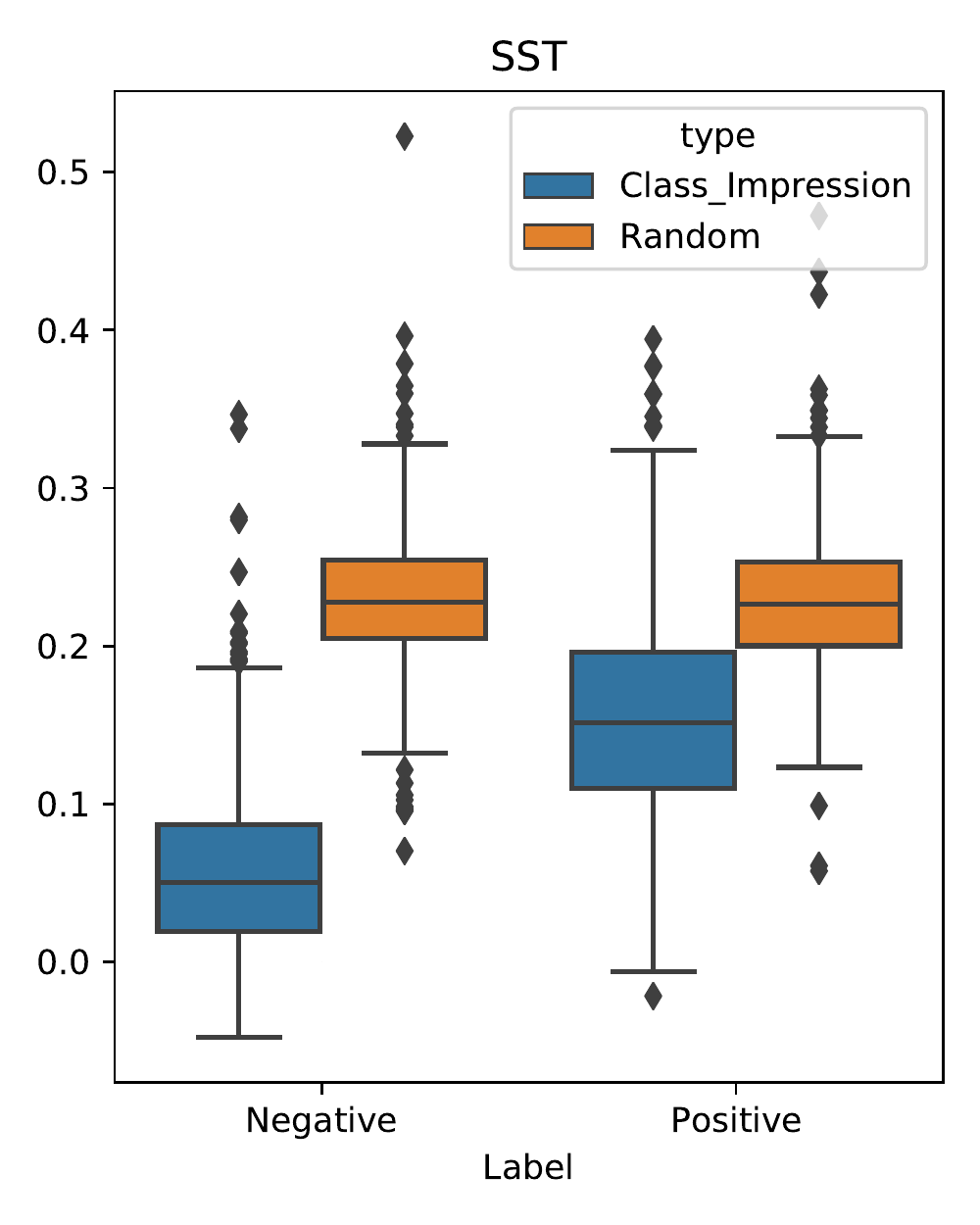} & \includegraphics[scale=0.35]{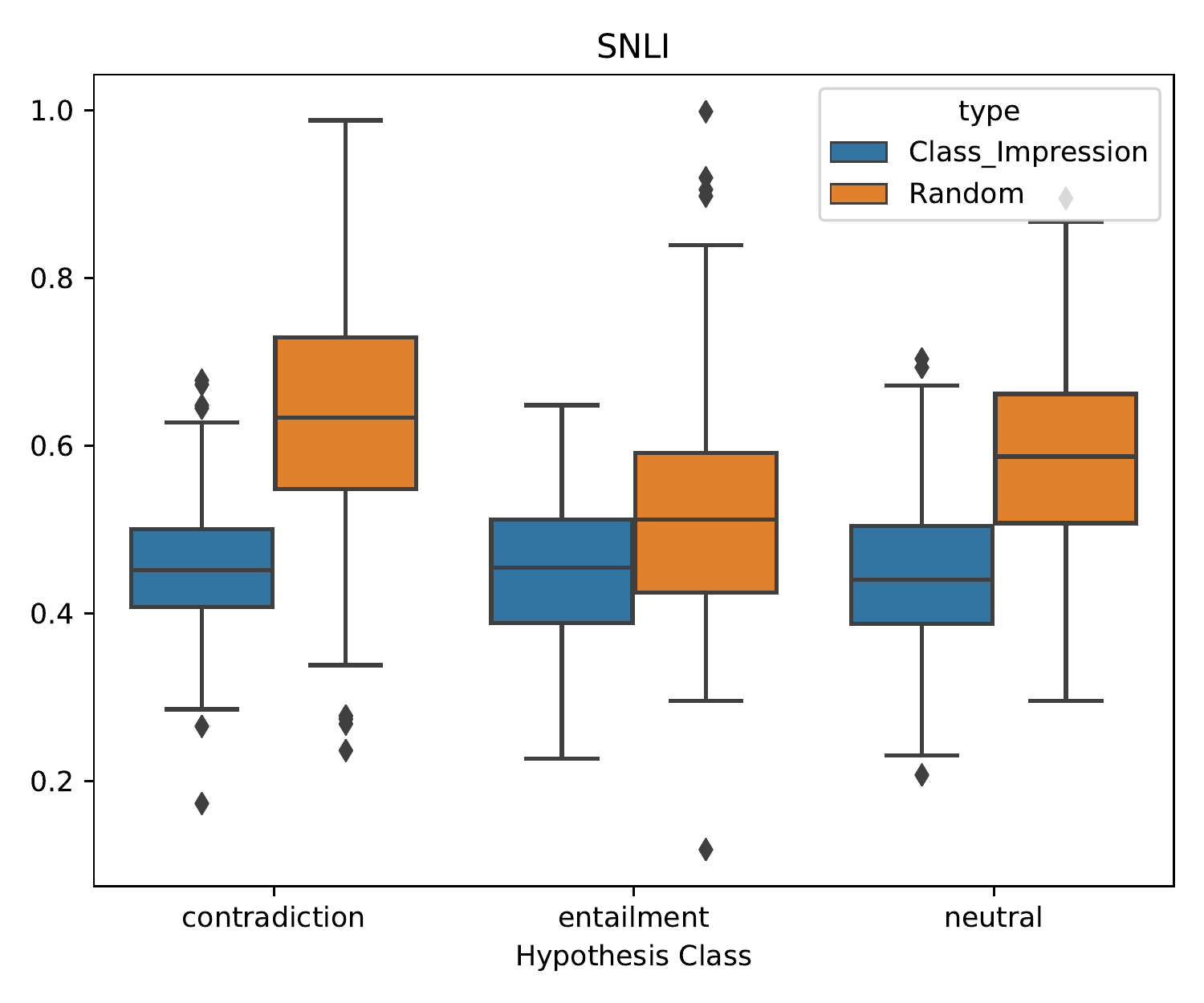} & \includegraphics[scale=0.35]{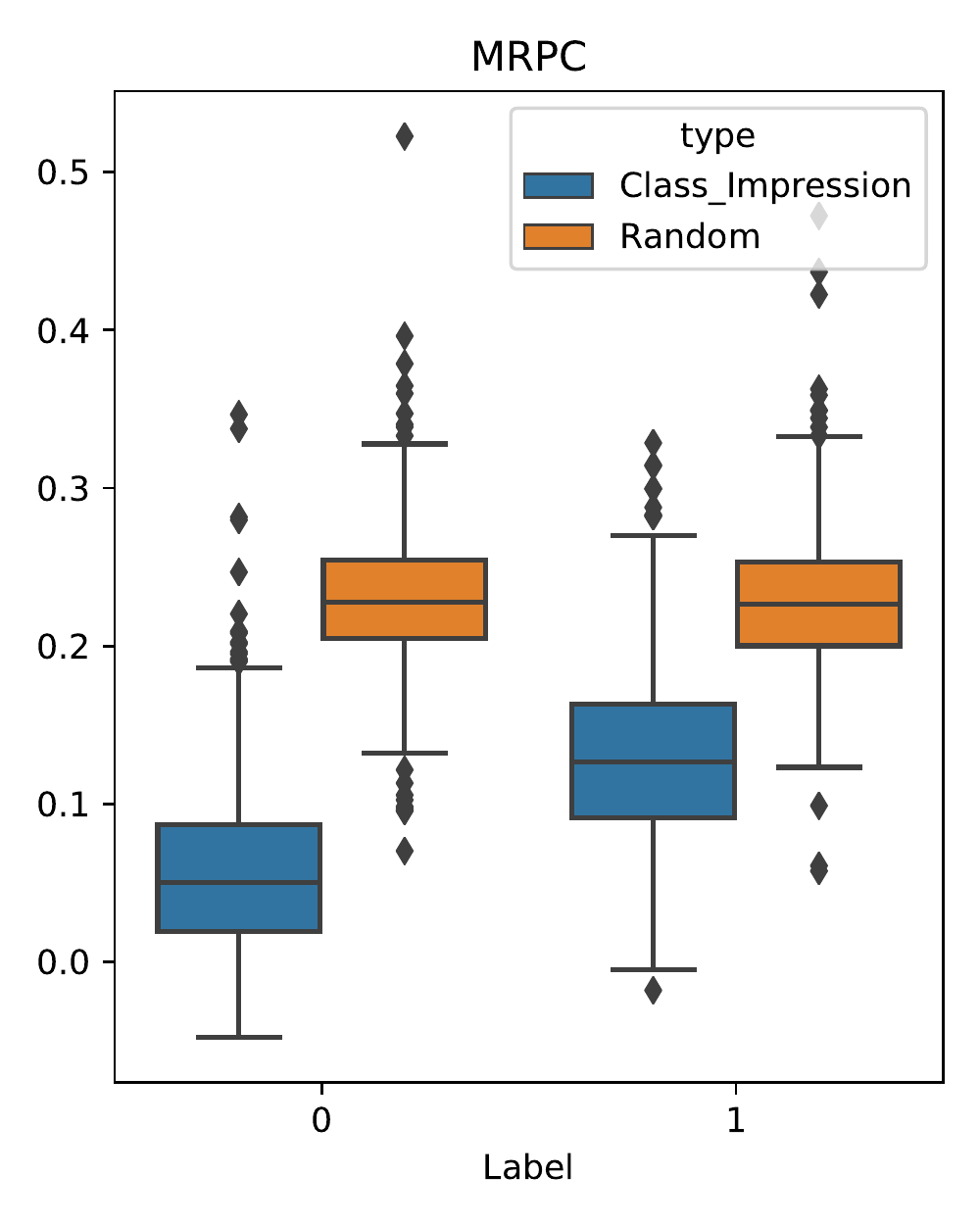}\\
 \end{tabular}

 \caption{\small 
 Mean Entropy of class impression words and 350 words randomly selected from the SST, SNLI, and MRPC dataset vocabularies. \label{fig:entropy-all-3-datasets}}


 \centering
 \begin{tabular}{ccc}
\includegraphics[scale=0.36]{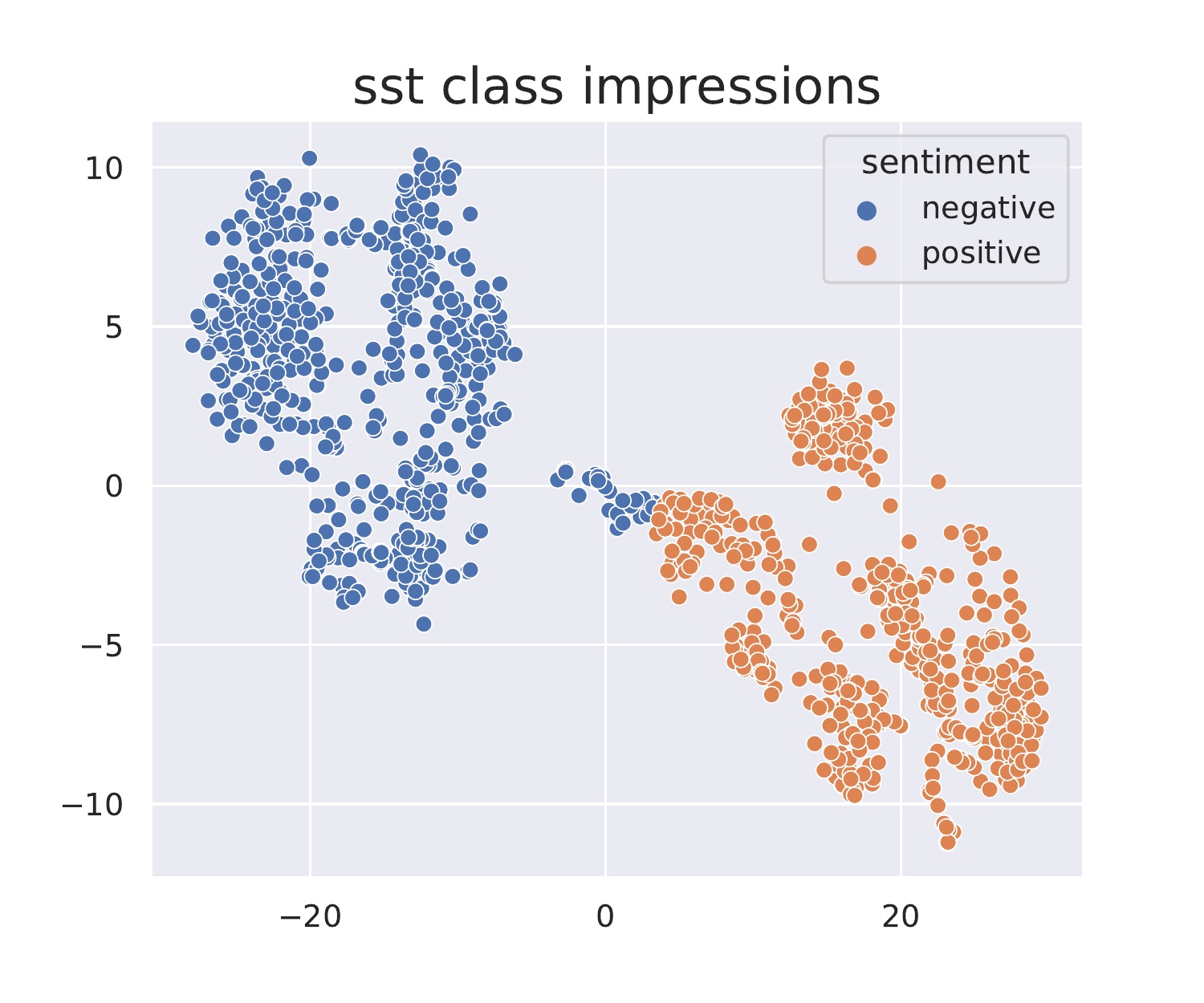}&\includegraphics[scale=0.36]{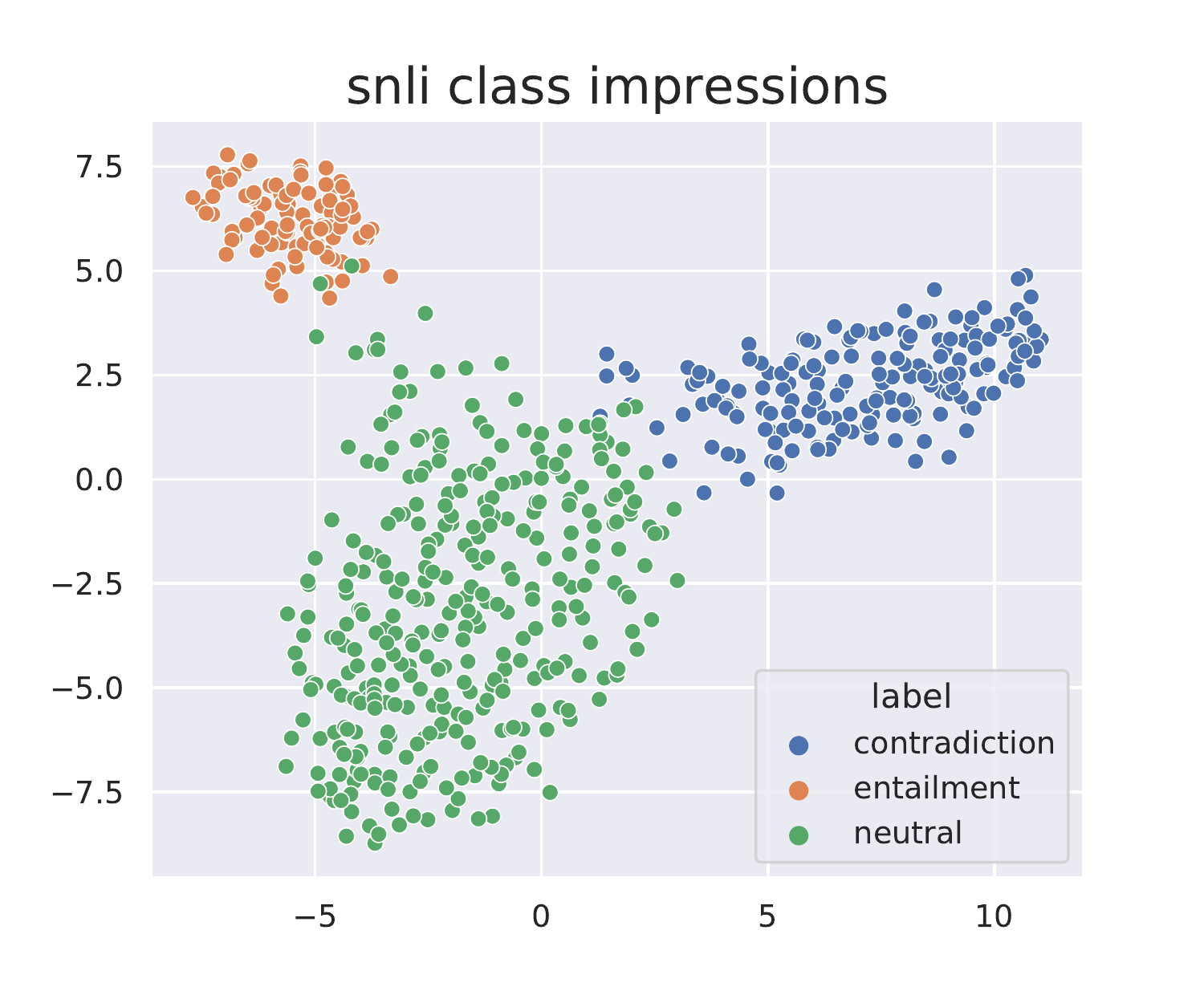}&\includegraphics[scale=0.36]{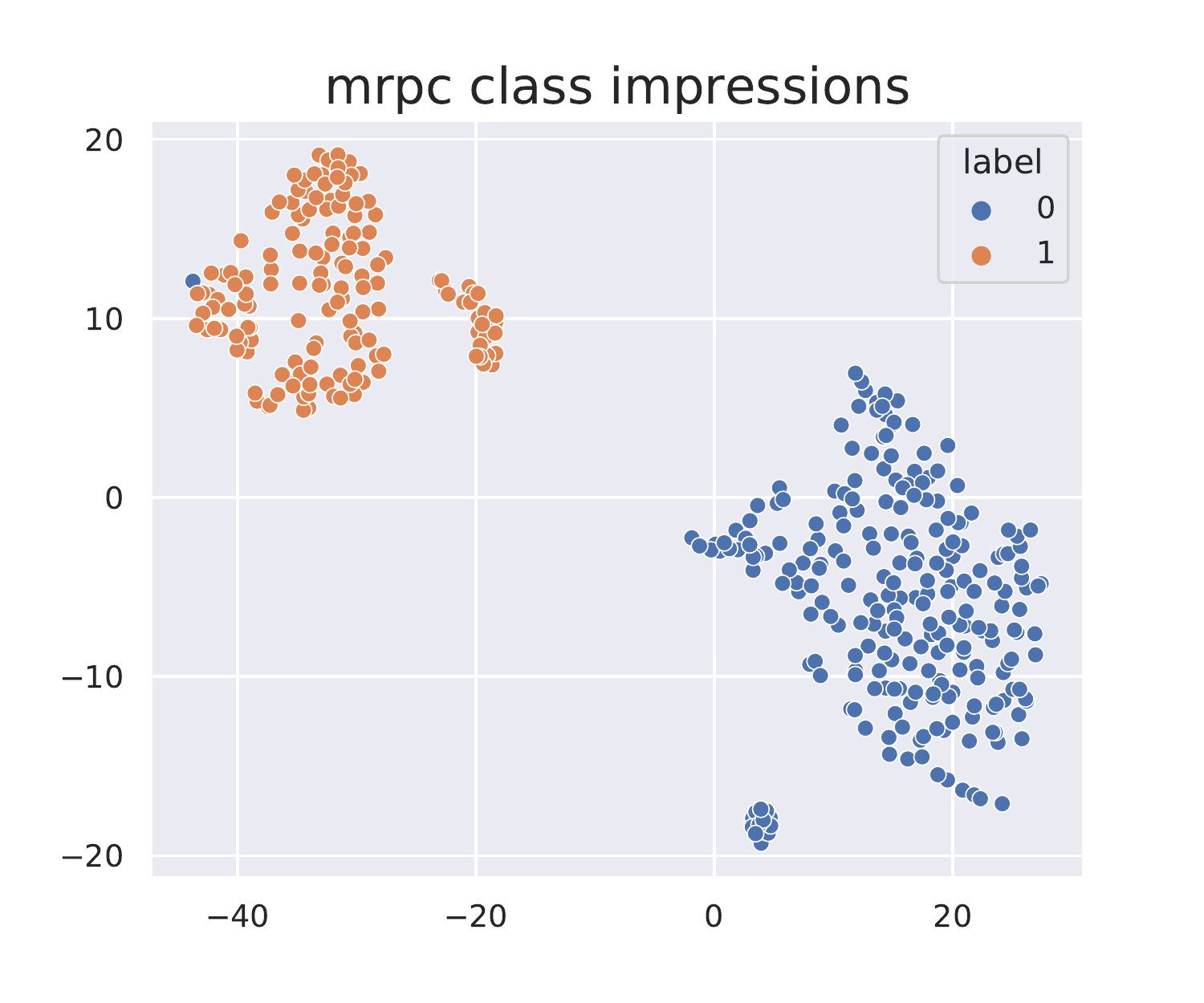}\\
 \end{tabular}

 \caption{\small 
 t-SNE plots for SST, SNLI, and MRPC class impression words. The words from the different class impressions form different distinct clusters depending on its class for all three datasets. The clusters are shown in different colors based on their classes. \label{fig:t-SNE plots}}

\end{figure*}

\textbf{Class Impression Words:} For analyzing why certain words are selected as representatives of a particular class, we find the discriminative power of each word by calculating its entropy. Concretely, we calculate entropy of the random variable $Y|X$ where $Y$ denotes a model class and $X$ denotes the word level feature. Formally, we compute:
\begin{equation}
    \mathbb{H}(Y|X) = -\sum_{k=1}^{K}p(Y=k|X)\log_2p(Y=k|X)
\end{equation}
for the class impression words and we compare them with randomly chosen words from the model vocabulary.
Fig.~\ref{fig:entropy-all-3-datasets} shows the results for SST, SNLI, and MRPC datasets. Interestingly, we find that the words which form class impressions are low entropy features. These words are much more discriminative than other randomly sampled words for all three datasets. This is further reinforced by Fig.~\ref{fig:t-SNE plots} where we show t-SNE plots for all the datasets \cy{do you use word-embedding feature?}. They show that words from different class impressions form distinct clusters.

Fig.~\ref{fig:entropy-all-3-datasets} shows that CIG algorithm selects low entropy features as representatives of different classes. However, it does not show the class-preference of these low entropy word-features. We hypothesize that those words become representatives of a particular class with a higher PMIs with respect to that class. In order to show this, we calculate PMI values of class representatives for each class and note that class representatives have a higher PMI for their own class than other classes. Formally, we compute:
\begin{equation}
    PMI(word, class) = log \frac{p(word,class)}{p(word,.)p(.,class)}
\end{equation}
We use add-10 smoothing for calculating this. We then group each class impression word based on its target class and report their PMI percentile. We show the results 
in Tables~\ref{table:sst-pmi-percentile}-\ref{table:snli-pmi-percentile}. It can be seen that class representatives have very high PMI percentiles. Previous studies have characterized high PMI words as \textit{dataset artefacts} \cite{gururangan2018annotation,poliak2018hypothesis}. \citet{wallace2019universal} have also shown that universal adversarial triggers have a high overlap with these dataset artefacts and consequently have high PMI values. Since we observe that class representatives too have high PMI values, we hypothesize that they could act as good adversarial triggers.

Following this, we postulate that adding class impression words of one class to a real example of another class should change the prediction of that example. For validating this, we conduct an experiment where we take words from class impressions of class $c_i$ and prepend them to real examples of class $c_j$. Table~\ref{table:snli-uat-ci-word-transfer} shows the results of the experiment over SNLI dataset. As can be seen, the results are very promising. 

We observe that the class which was more adversarially unsecure (\textit{Entailment}$\textgreater_{adv-unsecure}$\textit{Contradiction}) has better class impression words. These words, when added to examples of other classes, produce more successful perturbations. For e.g., when entailment words are added to contradiction examples, they reduce the accuracy from 91\% to less than 10\%. On the other hand, contradiction was adversarially more secure, and hence there is no appreciable reduction in the accuracy of any other class upon adding the contradiction class impression words\footnote{We find similar results on the MRPC dataset. We did not do these experiments for the SST dataset since SST class impression words are construct-relevant words and hence are bound to change sentiment scores while the same is not true for the other two datasets.}. This result can potentially help dataset designers design more secure datasets on which the model-makers can train adversarially robust models.

The above analysis shows that we can get class-impressions and adversarial triggers from dataset itself by computing entropy and PMI values. Moreover, our experiments in Sec.~\ref{sec:experiments} show that one can equivalently mine models to get class impressions and adversarial triggers. Therefore,
we conclude that we can craft both class impressions and adversarial triggers given either dataset or a well-trained model (\textit{i.e.}, the one which can model training data distribution well). Further, the models represent their classes with dataset artefacts. These artefacts are also responsible for making them adversarially unsecure. The lesser the dataset artefacts in a class, the lesser is a trained model's representative capacity for that class, and the more is the model's adversarial robustness for that class. We would like to further develop on these initial results to better dataset design protocols in future work.

\section{Conclusion and Future Work}
\label{sec:conclusion}
This paper presents a novel data-free approach, MINIMAL to mine natural language processing models for input-agnostic (universal) adversarial triggers. Our setting is more natural, which assumes an attacker does not have access to training data but only the trained model. Therefore, existing data-dependent adversarial trigger generation techniques are unrealistic in practice. On the other hand, our method is data-free and achieves comparable performance to data-based adversarial trigger generation methods. We also show that the triggers generated by our algorithm transfer remarkably well to different models and word embeddings. We achieve this by developing a combination of model inversion and adversarial trigger generation attacks. Finally, we show that low entropy word-level features occur as adversarial triggers and hence one can equivalently mine either a model or a dataset for these triggers.

We conduct our analysis on word-level triggers and class impressions based model inversion. While this analysis leads to crucial insights into dataset design and adversarial trigger crafting techniques, it can be extended to multi-word contextual analysis. 
This will also potentially lead to better dataset design protocols. We are actively engaged in this line of research. Further, another research focus can be to generate natural-looking class impressions and, consequently adversarial triggers.

{\small%
\bibliography{aaai22}}

\begin{thebibliography}{36}
\providecommand{\natexlab}[1]{#1}

\bibitem[{Behjati et~al.(2019)Behjati, Moosavi{-}Dezfooli, Baghshah, and
  Frossard}]{behjati2019universal}
Behjati, M.; Moosavi{-}Dezfooli, S.; Baghshah, M.~S.; and Frossard, P. 2019.
\newblock Universal Adversarial Attacks on Text Classifiers.
\newblock In \emph{{IEEE} International Conference on Acoustics, Speech and
  Signal Processing, {ICASSP} 2019, Brighton, United Kingdom, May 12-17, 2019},
  7345--7349. {IEEE}.

\bibitem[{Bowman et~al.(2015)Bowman, Angeli, Potts, and
  Manning}]{bowman2015large}
Bowman, S.~R.; Angeli, G.; Potts, C.; and Manning, C.~D. 2015.
\newblock A large annotated corpus for learning natural language inference.
\newblock In \emph{Proceedings of the 2015 Conference on Empirical Methods in
  Natural Language Processing}, 632--642. Lisbon, Portugal: Association for
  Computational Linguistics.

\bibitem[{Cambria et~al.(2013)Cambria, Schuller, Xia, and
  Havasi}]{cambria2013new}
Cambria, E.; Schuller, B.; Xia, Y.; and Havasi, C. 2013.
\newblock New avenues in opinion mining and sentiment analysis.
\newblock \emph{IEEE Intelligent systems}, 28(2): 15--21.

\bibitem[{Chen et~al.(2017)Chen, Zhu, Ling, Wei, Jiang, and
  Inkpen}]{chen2016enhanced}
Chen, Q.; Zhu, X.; Ling, Z.-H.; Wei, S.; Jiang, H.; and Inkpen, D. 2017.
\newblock Enhanced {LSTM} for Natural Language Inference.
\newblock In \emph{Proceedings of the 55th Annual Meeting of the Association
  for Computational Linguistics (Volume 1: Long Papers)}, 1657--1668.
  Vancouver, Canada: Association for Computational Linguistics.

\bibitem[{Devlin et~al.(2019)Devlin, Chang, Lee, and
  Toutanova}]{devlin2018bert}
Devlin, J.; Chang, M.-W.; Lee, K.; and Toutanova, K. 2019.
\newblock {BERT}: Pre-training of Deep Bidirectional Transformers for Language
  Understanding.
\newblock In \emph{Proceedings of the 2019 Conference of the North {A}merican
  Chapter of the Association for Computational Linguistics: Human Language
  Technologies, Volume 1 (Long and Short Papers)}, 4171--4186. Minneapolis,
  Minnesota: Association for Computational Linguistics.

\bibitem[{Dolan and Brockett(2005)}]{dolan2005automatically}
Dolan, W.~B.; and Brockett, C. 2005.
\newblock Automatically Constructing a Corpus of Sentential Paraphrases.
\newblock In \emph{Proceedings of the Third International Workshop on
  Paraphrasing ({IWP}2005)}.

\bibitem[{Ebrahimi et~al.(2018)Ebrahimi, Rao, Lowd, and
  Dou}]{ebrahimi2017hotflip}
Ebrahimi, J.; Rao, A.; Lowd, D.; and Dou, D. 2018.
\newblock {H}ot{F}lip: White-Box Adversarial Examples for Text Classification.
\newblock In \emph{Proceedings of the 56th Annual Meeting of the Association
  for Computational Linguistics (Volume 2: Short Papers)}, 31--36. Melbourne,
  Australia: Association for Computational Linguistics.

\bibitem[{Fredrikson, Jha, and Ristenpart(2015)}]{fredrikson2015model}
Fredrikson, M.; Jha, S.; and Ristenpart, T. 2015.
\newblock Model inversion attacks that exploit confidence information and basic
  countermeasures.
\newblock In \emph{Proceedings of the 22nd ACM SIGSAC Conference on Computer
  and Communications Security}, 1322--1333.

\bibitem[{Google(2021)}]{gcnlAPI}
Google. 2021.
\newblock The Google Natural Language API.
\newblock
  \url{https://cloud.google.com/natural-language#natural-language-api-demo}.

\bibitem[{Graves and Schmidhuber(2005)}]{graves2005framewise}
Graves, A.; and Schmidhuber, J. 2005.
\newblock Framewise phoneme classification with bidirectional LSTM and other
  neural network architectures.
\newblock \emph{Neural networks}, 18(5-6): 602--610.

\bibitem[{Gururangan et~al.(2018)Gururangan, Swayamdipta, Levy, Schwartz,
  Bowman, and Smith}]{gururangan2018annotation}
Gururangan, S.; Swayamdipta, S.; Levy, O.; Schwartz, R.; Bowman, S.; and Smith,
  N.~A. 2018.
\newblock Annotation Artifacts in Natural Language Inference Data.
\newblock In \emph{Proceedings of the 2018 Conference of the North {A}merican
  Chapter of the Association for Computational Linguistics: Human Language
  Technologies, Volume 2 (Short Papers)}, 107--112. New Orleans, Louisiana:
  Association for Computational Linguistics.

\bibitem[{Huan et~al.(2020)Huan, Wang, Zhang, Shang, Fu, and
  Zhou}]{huan2020data}
Huan, Z.; Wang, Y.; Zhang, X.; Shang, L.; Fu, C.; and Zhou, J. 2020.
\newblock Data-free adversarial perturbations for practical black-box attack.
\newblock In \emph{Pacific-Asia conference on knowledge discovery and data
  mining}, 127--138. Springer.

\bibitem[{Khrulkov and Oseledets(2018)}]{khrulkov2018art}
Khrulkov, V.; and Oseledets, I.~V. 2018.
\newblock Art of Singular Vectors and Universal Adversarial Perturbations.
\newblock In \emph{2018 {IEEE} Conference on Computer Vision and Pattern
  Recognition, {CVPR} 2018, Salt Lake City, UT, USA, June 18-22, 2018},
  8562--8570. {IEEE} Computer Society.

\bibitem[{Kumar et~al.(2019)Kumar, Aggarwal, Mahata, Shah, Kumaraguru, and
  Zimmermann}]{kumar2019get}
Kumar, Y.; Aggarwal, S.; Mahata, D.; Shah, R.~R.; Kumaraguru, P.; and
  Zimmermann, R. 2019.
\newblock Get {IT} Scored Using AutoSAS - An Automated System for Scoring Short
  Answers.
\newblock In \emph{The Thirty-Third {AAAI} Conference on Artificial
  Intelligence, {AAAI} 2019, The Thirty-First Innovative Applications of
  Artificial Intelligence Conference, {IAAI} 2019, The Ninth {AAAI} Symposium
  on Educational Advances in Artificial Intelligence, {EAAI} 2019, Honolulu,
  Hawaii, USA, January 27 - February 1, 2019}, 9662--9669. {AAAI} Press.

\bibitem[{Lan et~al.(2020)Lan, Chen, Goodman, Gimpel, Sharma, and
  Soricut}]{lan2019albert}
Lan, Z.; Chen, M.; Goodman, S.; Gimpel, K.; Sharma, P.; and Soricut, R. 2020.
\newblock {ALBERT:} {A} Lite {BERT} for Self-supervised Learning of Language
  Representations.
\newblock In \emph{8th International Conference on Learning Representations,
  {ICLR} 2020, Addis Ababa, Ethiopia, April 26-30, 2020}. OpenReview.net.

\bibitem[{Li et~al.(2019)Li, Ji, Liu, Hong, Gao, and Tian}]{li2019universal}
Li, J.; Ji, R.; Liu, H.; Hong, X.; Gao, Y.; and Tian, Q. 2019.
\newblock Universal Perturbation Attack Against Image Retrieval.
\newblock In \emph{2019 {IEEE/CVF} International Conference on Computer Vision,
  {ICCV} 2019, Seoul, Korea (South), October 27 - November 2, 2019},
  4898--4907. {IEEE}.

\bibitem[{Meng et~al.(2017)Meng, Zhao, Han, He, Brusilovsky, and
  Chi}]{meng2017deep}
Meng, R.; Zhao, S.; Han, S.; He, D.; Brusilovsky, P.; and Chi, Y. 2017.
\newblock Deep Keyphrase Generation.
\newblock In \emph{Proceedings of the 55th Annual Meeting of the Association
  for Computational Linguistics (Volume 1: Long Papers)}, 582--592. Vancouver,
  Canada: Association for Computational Linguistics.

\bibitem[{Micaelli and Storkey(2019)}]{micaelli2019zero}
Micaelli, P.; and Storkey, A.~J. 2019.
\newblock Zero-shot Knowledge Transfer via Adversarial Belief Matching.
\newblock In Wallach, H.~M.; Larochelle, H.; Beygelzimer, A.;
  d'Alch{\'{e}}{-}Buc, F.; Fox, E.~B.; and Garnett, R., eds., \emph{Advances in
  Neural Information Processing Systems 32: Annual Conference on Neural
  Information Processing Systems 2019, NeurIPS 2019, December 8-14, 2019,
  Vancouver, BC, Canada}, 9547--9557.

\bibitem[{Michel et~al.(2019)Michel, Li, Neubig, and
  Pino}]{michel2019evaluation}
Michel, P.; Li, X.; Neubig, G.; and Pino, J. 2019.
\newblock On Evaluation of Adversarial Perturbations for Sequence-to-Sequence
  Models.
\newblock In \emph{Proceedings of the 2019 Conference of the North {A}merican
  Chapter of the Association for Computational Linguistics: Human Language
  Technologies, Volume 1 (Long and Short Papers)}, 3103--3114. Minneapolis,
  Minnesota: Association for Computational Linguistics.

\bibitem[{Mikolov et~al.(2018)Mikolov, Grave, Bojanowski, Puhrsch, and
  Joulin}]{mikolov2017advances}
Mikolov, T.; Grave, E.; Bojanowski, P.; Puhrsch, C.; and Joulin, A. 2018.
\newblock Advances in Pre-Training Distributed Word Representations.
\newblock In \emph{Proceedings of the Eleventh International Conference on
  Language Resources and Evaluation ({LREC} 2018)}. Miyazaki, Japan: European
  Language Resources Association (ELRA).

\bibitem[{Moosavi{-}Dezfooli et~al.(2017)Moosavi{-}Dezfooli, Fawzi, Fawzi, and
  Frossard}]{moosavi2017universal}
Moosavi{-}Dezfooli, S.; Fawzi, A.; Fawzi, O.; and Frossard, P. 2017.
\newblock Universal Adversarial Perturbations.
\newblock In \emph{2017 {IEEE} Conference on Computer Vision and Pattern
  Recognition, {CVPR} 2017, Honolulu, HI, USA, July 21-26, 2017}, 86--94.
  {IEEE} Computer Society.

\bibitem[{Mopuri, Ganeshan, and Babu(2018)}]{mopuri2018generalizable}
Mopuri, K.~R.; Ganeshan, A.; and Babu, R.~V. 2018.
\newblock Generalizable data-free objective for crafting universal adversarial
  perturbations.
\newblock \emph{IEEE transactions on pattern analysis and machine
  intelligence}, 41(10): 2452--2465.

\bibitem[{Mopuri, Garg, and Radhakrishnan(2017)}]{mopuri2017fast}
Mopuri, K.~R.; Garg, U.; and Radhakrishnan, V.~B. 2017.
\newblock Fast Feature Fool: {A} data independent approach to universal
  adversarial perturbations.
\newblock In \emph{British Machine Vision Conference 2017, {BMVC} 2017, London,
  UK, September 4-7, 2017}. {BMVA} Press.

\bibitem[{Mopuri, Uppala, and Babu(2018)}]{mopuri2018ask}
Mopuri, K.~R.; Uppala, P.~K.; and Babu, R.~V. 2018.
\newblock Ask, acquire, and attack: Data-free uap generation using class
  impressions.
\newblock In \emph{Proceedings of the European Conference on Computer Vision
  (ECCV)}, 19--34.

\bibitem[{Nayak et~al.(2019)Nayak, Mopuri, Shaj, Radhakrishnan, and
  Chakraborty}]{nayak2019zero}
Nayak, G.~K.; Mopuri, K.~R.; Shaj, V.; Radhakrishnan, V.~B.; and Chakraborty,
  A. 2019.
\newblock Zero-Shot Knowledge Distillation in Deep Networks.
\newblock In Chaudhuri, K.; and Salakhutdinov, R., eds., \emph{Proceedings of
  the 36th International Conference on Machine Learning, {ICML} 2019, 9-15 June
  2019, Long Beach, California, {USA}}, volume~97 of \emph{Proceedings of
  Machine Learning Research}, 4743--4751. {PMLR}.

\bibitem[{Parikh et~al.(2016)Parikh, T{\"a}ckstr{\"o}m, Das, and
  Uszkoreit}]{parikh2016decomposable}
Parikh, A.; T{\"a}ckstr{\"o}m, O.; Das, D.; and Uszkoreit, J. 2016.
\newblock A Decomposable Attention Model for Natural Language Inference.
\newblock In \emph{Proceedings of the 2016 Conference on Empirical Methods in
  Natural Language Processing}, 2249--2255. Austin, Texas: Association for
  Computational Linguistics.

\bibitem[{Pennington, Socher, and Manning(2014)}]{pennington2014glove}
Pennington, J.; Socher, R.; and Manning, C. 2014.
\newblock {G}lo{V}e: Global Vectors for Word Representation.
\newblock In \emph{Proceedings of the 2014 Conference on Empirical Methods in
  Natural Language Processing ({EMNLP})}, 1532--1543. Doha, Qatar: Association
  for Computational Linguistics.

\bibitem[{Peters et~al.(2018)Peters, Neumann, Iyyer, Gardner, Clark, Lee, and
  Zettlemoyer}]{peters2018deep}
Peters, M.; Neumann, M.; Iyyer, M.; Gardner, M.; Clark, C.; Lee, K.; and
  Zettlemoyer, L. 2018.
\newblock Deep Contextualized Word Representations.
\newblock In \emph{Proceedings of the 2018 Conference of the North {A}merican
  Chapter of the Association for Computational Linguistics: Human Language
  Technologies, Volume 1 (Long Papers)}, 2227--2237. New Orleans, Louisiana:
  Association for Computational Linguistics.

\bibitem[{Poliak et~al.(2018)Poliak, Naradowsky, Haldar, Rudinger, and
  Van~Durme}]{poliak2018hypothesis}
Poliak, A.; Naradowsky, J.; Haldar, A.; Rudinger, R.; and Van~Durme, B. 2018.
\newblock Hypothesis Only Baselines in Natural Language Inference.
\newblock In \emph{Proceedings of the Seventh Joint Conference on Lexical and
  Computational Semantics}, 180--191. New Orleans, Louisiana: Association for
  Computational Linguistics.

\bibitem[{Socher et~al.(2013)Socher, Perelygin, Wu, Chuang, Manning, Ng, and
  Potts}]{socher2013recursive}
Socher, R.; Perelygin, A.; Wu, J.; Chuang, J.; Manning, C.~D.; Ng, A.; and
  Potts, C. 2013.
\newblock Recursive Deep Models for Semantic Compositionality Over a Sentiment
  Treebank.
\newblock In \emph{Proceedings of the 2013 Conference on Empirical Methods in
  Natural Language Processing}, 1631--1642. Seattle, Washington, USA:
  Association for Computational Linguistics.

\bibitem[{Song et~al.(2021)Song, Yu, Peng, and Narasimhan}]{song2020universal}
Song, L.; Yu, X.; Peng, H.-T.; and Narasimhan, K. 2021.
\newblock Universal Adversarial Attacks with Natural Triggers for Text
  Classification.
\newblock In \emph{Proceedings of the 2021 Conference of the North American
  Chapter of the Association for Computational Linguistics: Human Language
  Technologies}, 3724--3733. Online: Association for Computational Linguistics.

\bibitem[{Tram{\`e}r et~al.(2016)Tram{\`e}r, Zhang, Juels, Reiter, and
  Ristenpart}]{tramer2016stealing}
Tram{\`e}r, F.; Zhang, F.; Juels, A.; Reiter, M.~K.; and Ristenpart, T. 2016.
\newblock Stealing machine learning models via prediction apis.
\newblock In \emph{25th $\{$USENIX$\}$ Security Symposium ($\{$USENIX$\}$
  Security 16)}, 601--618.

\bibitem[{Wallace et~al.(2019)Wallace, Feng, Kandpal, Gardner, and
  Singh}]{wallace2019universal}
Wallace, E.; Feng, S.; Kandpal, N.; Gardner, M.; and Singh, S. 2019.
\newblock Universal Adversarial Triggers for Attacking and Analyzing {NLP}.
\newblock In \emph{Proceedings of the 2019 Conference on Empirical Methods in
  Natural Language Processing and the 9th International Joint Conference on
  Natural Language Processing (EMNLP-IJCNLP)}, 2153--2162. Hong Kong, China:
  Association for Computational Linguistics.

\bibitem[{Xiong, Zhong, and Socher(2017)}]{xiong2016dynamic}
Xiong, C.; Zhong, V.; and Socher, R. 2017.
\newblock Dynamic Coattention Networks For Question Answering.
\newblock In \emph{5th International Conference on Learning Representations,
  {ICLR} 2017, Toulon, France, April 24-26, 2017, Conference Track
  Proceedings}. OpenReview.net.

\bibitem[{Zhang et~al.(2021)Zhang, Benz, Lin, Karjauv, Wu, and
  Kweon}]{zhang2021survey}
Zhang, C.; Benz, P.; Lin, C.; Karjauv, A.; Wu, J.; and Kweon, I.~S. 2021.
\newblock A survey on universal adversarial attack.
\newblock \emph{arXiv preprint arXiv:2103.01498}.

\bibitem[{Zhang, Wang, and Liu(2018)}]{zhang2018deep}
Zhang, L.; Wang, S.; and Liu, B. 2018.
\newblock Deep learning for sentiment analysis: A survey.
\newblock \emph{Wiley Interdisciplinary Reviews: Data Mining and Knowledge
  Discovery}, 8(4): e1253.

\end{thebibliography}

\end{document}